\documentclass[conference]{IEEEtran}
\IEEEoverridecommandlockouts
\usepackage{cite}
\usepackage{amsmath,amssymb,amsfonts}
\usepackage{algorithmic}
\usepackage{graphicx}
\usepackage{textcomp}
\usepackage{xcolor}
\usepackage{amsthm,amsmath,amssymb}
\usepackage{mathrsfs}
\usepackage{bm}
\usepackage{CJKutf8}
\graphicspath{{Fig/}}
\def\BibTeX{{\rm B\kern-.05em{\sc i\kern-.025em b}\kern-.08em
    T\kern-.1667em\lower.7ex\hbox{E}\kern-.125emX}}
\begin{document}

\title{Significant Wave Height Prediction based on Wavelet Graph Neural Network
\thanks{This work was partially funded by Natural Science Foundation of Jiangsu Province under Grant No. BK20191298, Fundamental Research Funds for the Central Universities under Grant No. B200202175.}
}

\author{\IEEEauthorblockN{
    \begin{CJK}{UTF8}{gbsn}
        Delong Chen\textsuperscript{1},
    \end{CJK} 
    Fan Liu\textsuperscript{1}, 
    Zheqi Zhang\textsuperscript{2,}*, 
    Xiaomin Lu\textsuperscript{1}, 
    Zewen Li\textsuperscript{1} 
}
\IEEEauthorblockA{
\textsuperscript{1}College of Computer and Information Hohai University, Nanjing, China\\
\textsuperscript{2}Mihaylo College of Business and Economics, 
California State University, Fullerton, USA\\
email: tedzhang1988@gmail.com}

}

\maketitle

\begin{abstract}
    Computational intelligence-based ocean characteristics forecasting applications, such as Significant Wave Height (SWH) prediction, are crucial for avoiding social and economic loss in coastal cities. Compared to the traditional empirical-based or numerical-based forecasting models, ``soft computing'' approaches, including machine learning and deep learning models, have shown numerous success in recent years. In this paper, we focus on enabling the deep learning model to learn both short-term and long-term spatial-temporal dependencies for SWH prediction. A Wavelet Graph Neural Network (WGNN) approach is proposed to integrate the advantages of wavelet transform and graph neural network. Several parallel graph neural networks are separately trained on wavelet decomposed data, and the reconstruction of each model's prediction forms the final SWH prediction. Experimental results show that the proposed WGNN approach outperforms other models, including the numerical models, the machine learning models, and several deep learning models.
\end{abstract}

\begin{IEEEkeywords}
Significant wave height, Wavelet transform, Deep learning
\end{IEEEkeywords}

\section{Introduction}
   Ocean waves with high Significant Wave Height (SWH, or $H_s$) can overturn ships and destroy ocean or coastal engineering. It threatens human life, crop production, and the survival of aquaculture products. Therefore, the accurate prediction of SWH is vital since it can help reduce social and commercial losses. Moreover, SWH prediction can also bring several benefits. For example, optimizing ship routes according to the SWH prediction can avoid rough sea areas, thereby reducing the sailing time and fuel expenses. Furthermore, SWH prediction can provide valuable information for planning military and amphibious operations.
   
   Due to its importance and valuable applications, SWH prediction approaches have been continuously developed for decades. The empirical-based and numerical-based SWH prediction approaches in the early years have high interpretability but low accuracy and limited generalization ability. As the rise of computational intelligence, machine learning-based SWH prediction models, such as the Support Vector Machine (SVM) and the Artificial Neural Network (ANN), have shown their advantages. Especially in recent years, deep learning-based models, which hold strong feature extraction ability, have also been applied to SWH prediction successfully \cite{Quach2021Deep, Guan2020Wave}.
   
\begin{figure}
    \centering
    \includegraphics[width=0.46\textwidth]{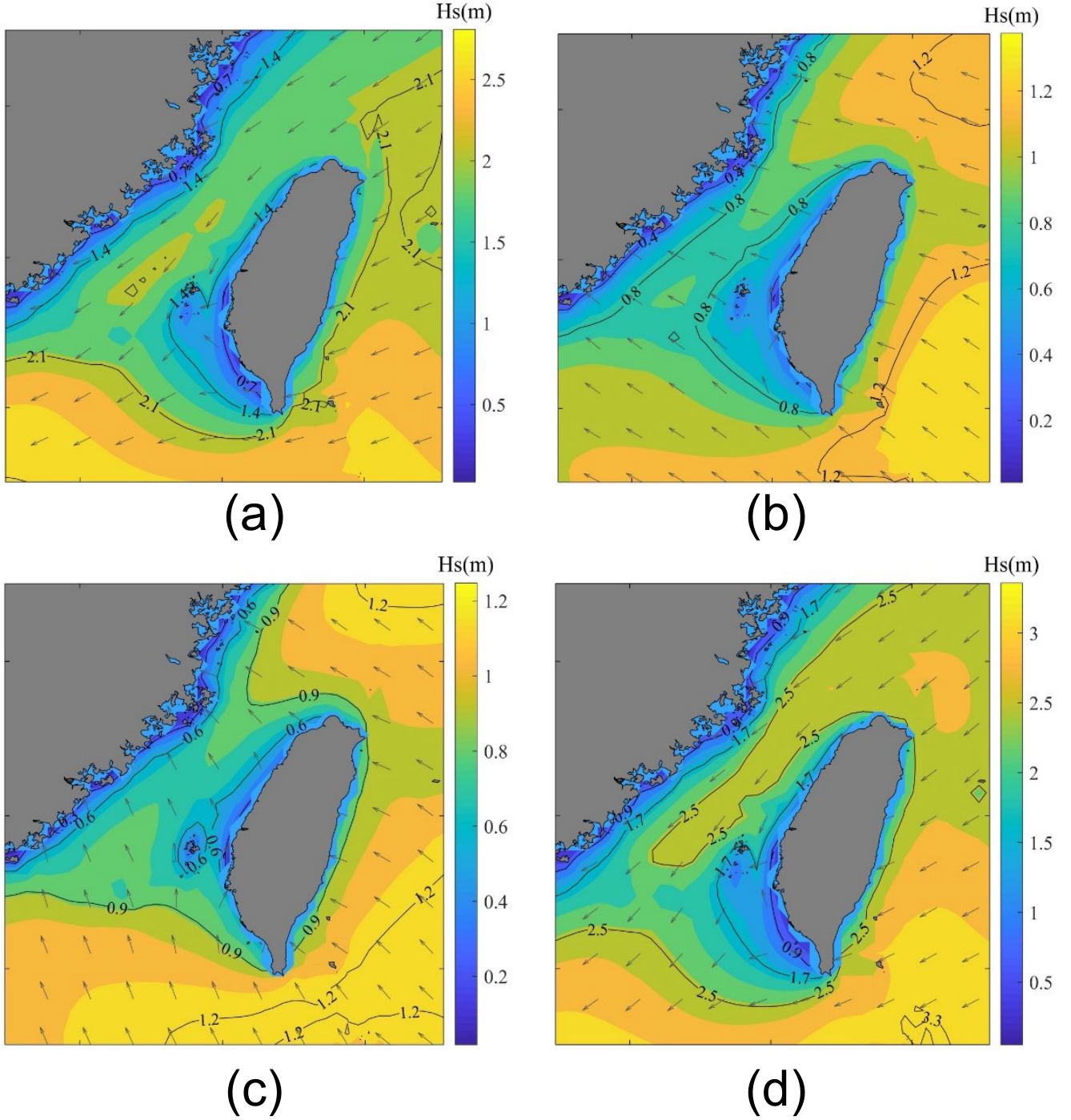}
    \caption{The seasonal variation of SWH caused by the monsoon. (a)-(d) respectively show the average SWH value and wave direction of Q1-Q4. This figure clarified the motivation and necessity of wavelet decomposition for SWH prediction.}
    \label{fig:Heatmap}
\end{figure}
   
   However, by reviewing the existing approaches, we find that there are the following two challenges that still remain and need to be solved for SWH prediction: 
   \textbf{1)} effectively capture the relationships between different types of inputs and learn its complicated non-linear mapping and temporal dependencies with the SWH data, and
   \textbf{2)} distinguish occasional extreme sea conditions and seasonal SWH variation and learn both short-term and long-term SWH patterns (see Fig.\ref{fig:Heatmap} for an example of long-term SWH variation). 
   
   In this paper, the above issues are addressed by the proposed \textbf{W}avelet \textbf{G}raph \textbf{N}eural \textbf{N}etwork (WGNN). The inputs and the target outputs are decomposed by the Debauches (Db)-type mother wavelet-based wavelet transform. For the derived components, several Graph Neural Networks (GNN) are separately deployed to learn the data dependencies in corresponding frequency band individually. The GNN can effectively capture the spatial-temporal pattern of data, especially the relationship between different types of inputs. Finally, the outputs of each GNN are reconstructed to form the final SWH prediction. To our best knowledge, it is the first time that a GNN is applied to the SWH prediction task, and also the first time that deep learning is integrated with wavelet transform for the SWH prediction. In the experiment, the effectiveness of the proposed WGNN is well validated. We compare WGNN with several SWH prediction baselines, and we find WGNN achieve the best performance. 
   
   The rest of this paper is organized as follows. In Section~\ref{sec:relatedworks}, we review existing SWH prediction methods, including machine learning-based, deep learning-based, and wavelet transform-based approaches. In Section~\ref{sec:approach}, we present the technical details of our proposed WGNN approach. Section~\ref{sec:experiments} and Section~\ref{sec:conclusion} compares different methods to make SWH predictions and summarizes findings respectively.

    \begin{figure*}[h]
        \centering
        \includegraphics[width=0.75\textwidth]{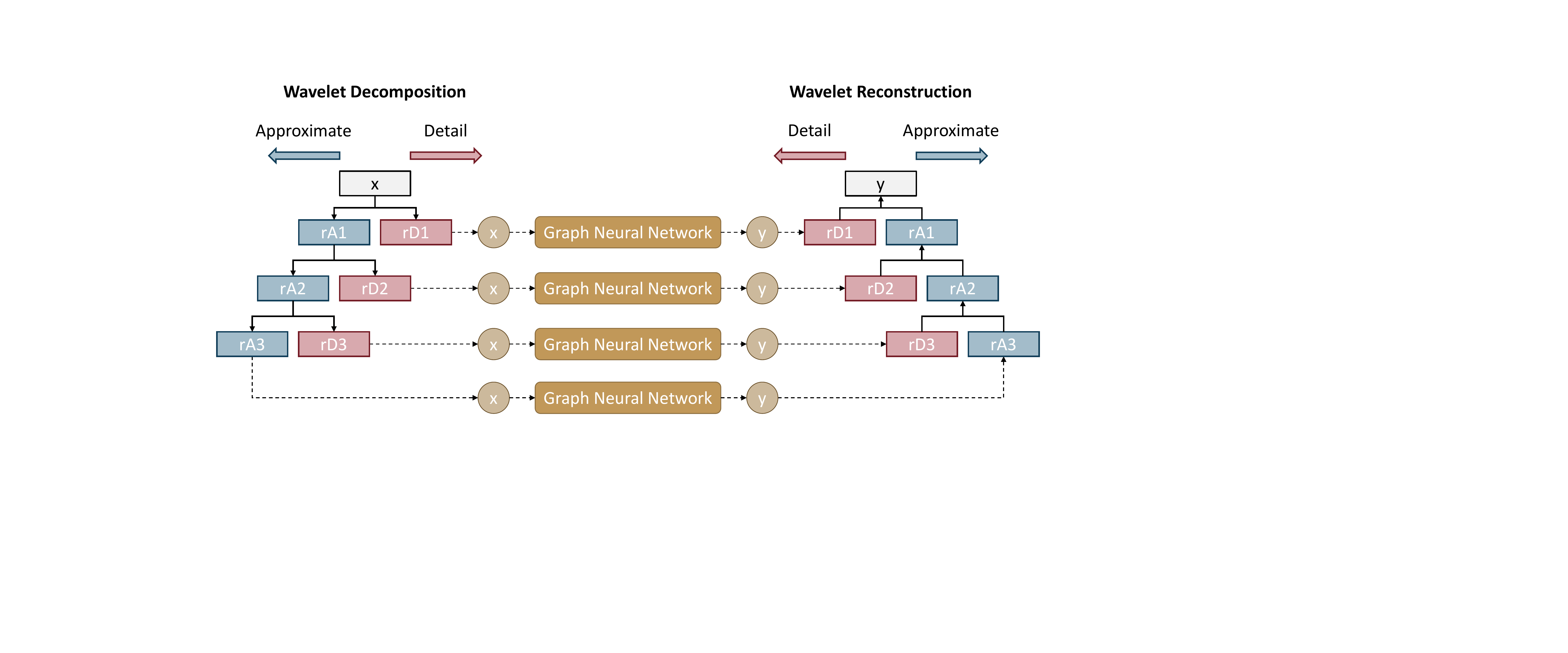}
        \caption{The illustration of the proposed WGNN.}
        \label{fig:WGNN}
    \end{figure*}

\section{Related Works}
\label{sec:relatedworks}
\subsection{Machine Learning-based Approaches}
    In last century, due to the difficulty of data collection and the limitation of computing power, SWH prediction was mainly based on empirical or numerical models \cite{Wamdi1988WAM, Booij1996SWAN, Mentaschi2015Performance}. Due to the lack of learning ability, these approaches have strong interpretability but low prediction accuracy and weak generalize performance. As the rapid development of machine learning theory, many machine learning algorithms, such as Support Vector Regression (SVR), Bayesian Network, XGBoost, extreme learning machine, and ANN, have been successfully deployed on the SWH prediction \cite{Mahjoobi2009Prediction, Malekmohamadi2011Evaluating, Deo1998Real}. These approaches were also named ``soft computing'' as the contrast of previous empirical-based or numerical-based ``hard computing'' approaches.

    It is worth noting that among these ``soft computing'' methods, ANN is the most widely used learning model \cite{Deo1998Real, Rizianiza2015Prediction, Berbic2017Application, James2018Machine}. Since ANN has the ability to establish complicated non-linear mapping to fit arbitrary functions, it has been widely valued and became a popular choice in SWH prediction. Malekmohamadi et al. \cite{Malekmohamadi2011Evaluating} evaluated SVM, Bayesian Network, ANN, and Adaptive Neuro-Fuzzy Inference System for SWH prediction and found that ANN achieved the best performance. In recent years, several variances of ANN, such as General Regression Neural Network \cite{Juliani2020Wave}, and various optimizers of ANN, such as cuckoo search algorithm \cite{Yang2019Prediction}, mind evolutionary algorithm \cite{Wang2018BP} have been investigated to improve the accuracy of SWH prediction.

    However, as a common drawback of the above models, the temporal dependencies of data are neglected, making the model sensitive to noise and therefore limiting their reliability. To solve this issue, Recurrent Neural Network (RNN)-based approach \cite{Mandal2006Ocean} and Long Short Term Memory (LSTM)-based approach \cite{Osawa2020Wave} have been developed. Many researchers have proved the advantages of RNN and LSTM through comparative experiments. For example, Abhigna et al. \cite{Abhigna2017Analysis} compared ANN and RNN and found that RNN could achieve lower error, and Gao et al. \cite{Gao2021Forcasting} showed the performance of LSTM is significantly better than SWAN \cite{Booij1996SWAN}, ANN and SVR.

\subsection{Deep Learning-based Approaches}
    A noticing research trend of SWH prediction is the usage of deep learning models. Compared to the above machine learning, deep learning has a stronger feature extraction capability. Zhang et al. \cite{Zhang2019Significant} used Conditional Restricted Boltzmann Machine-based Deep Belief Network (CRBF-DBN) to predict SWH. Kumar et al. \cite{Kumar2018Ocean} transferred a pre-trained Deep Belief Networks (DBN) to predict SWH of different locations to solve the issue of lacking sufficient training data. It demonstrates that the model could effectively learn generalized data patterns and dependencies from the pre-training dataset. In recent years, Convolutional Neural Network (CNN) \cite{LiZewenReview} was also found to have great potential. For example, Quach et al. \cite{Quach2021Deep} used a CNN to extract information from Synthetic Aperture Radar (SAR) images and predict SWH. Compared to previous works, their deep learning-based approach achieved significantly better performance. This success demonstrated the feasibility and effectiveness of convolutional deep models for the SWH prediction. 
    
    Choi et al. \cite{Choi2020Real} used a CNN-LSTM model to predict SWH from solely raw ocean image, but this method fails in foggy or night conditions. The CNN was originally designed for image-oriented tasks \cite{Krizhevsky2012ImageNet}, and just as in the above methods, CNN is used to extract visual features. Despite visual data, researchers also found CNN's feasibility on processing raw sensor signal data. For example, recently Guan et al. \cite{Guan2020Wave} proposed to use a CNN-LSTM model for SWH prediction. Their model outperforms the model based on solely LSTM, showing that CNN can effectively extract local features from sensor signal data and improve the performance. Besides, Wang et al. and Mooneyham et al. combined the advantages of ``hard computing'' numerical models and ``soft computing'' learning-based models by using a multi-factor extreme learning machine \cite{Wang2021Residual} and CNN \cite{Mooneyham2020SWRL} to correct the prediction of numerical model, which is also an emerging and promising approach for SWH prediction.

\subsection{Wavelet Transform-based Approaches}
    The SWH data, as well as the inputs of the SWH prediction model, are non-stationary time series composing of different components with different periods and frequencies. These different components are affected by different factors, such as tides, weather, monsoon, seasons, etc. Researchers found separately deploying several models to learn different components could lead to higher accuracy. For the SWH prediction, wavelet transform has been widely used to perform decomposition and be combined with a variety of machine learning methods, such as fussy logic \cite{Oezger2010Significant}, and genetic programming \cite{Shahabi2017Significant}. The combination of wavelet transform and ANN has received a lot of attention. For example, Deshmukh et al. \cite{Deshmukh2016Neural} decomposed the residual of the numerical model and ground truth data by wavelets, then they trained an ANN to correct the numerical prediction. Dixit et al. \cite{Dixit2016Review} named the combination of wavelet transform as  Neuro-Wavelet Techniques and presented a review of its hydrology and ocean applications. However, to our best knowledge, existing wavelet transform-based approaches only use shallow networks, but whether a deep model can be successfully combined with wavelet transform is still unclear.

\section{Approach}
\label{sec:approach}

    \begin{figure*}[h]
        \centering
        \includegraphics[width=0.98\textwidth]{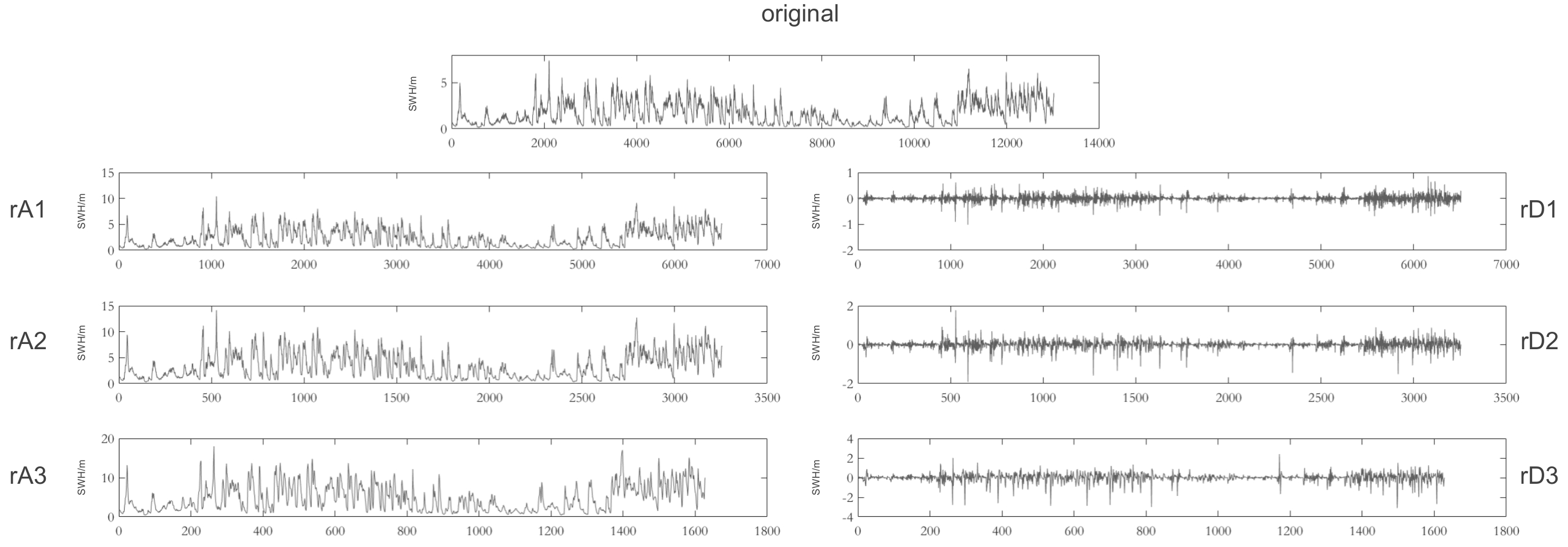}
        \caption{An example of SWH wavelet decomposition.}
        \label{fig:Wavelet123}
    \end{figure*}

    Suppose we have a dataset $\mathcal{D}=\{(\bm{X}_i, \bm{Y}_i)\}_{i=1}^N$, where $\bm{X}_i=\{\bm{x}_t\}_{t=1}^T$ and $\bm{Y}_i=\{\bm{y}_t\}_{t=1}^T$ are respectively the input SWH-related signal data sequences and ground truth SWH sequences. Each $\bm{x}_t$ is a three-dimensional vector containing the information of average wind speed, maximum wind speed and wind direction at time step $t$, and $\bm{y}_t$ is the ground truth SWH value. Our goal is to estimate a function $f$ from $\mathcal{D}$ to predict SWH from given signal data, i.e. $\hat{\bm{Y}}_i=f(\bm{X}_i)$. Our proposed approach is based on wavelet decomposition and reconstruction. As illustrated in Fig.\ref{fig:WGNN}, we first decompose $X_i$ and $Y_i$ into a series of components, i.e. $\bm{X}_i = \bm{X}_i^1 + \bm{X}_i^2 + ... + \bm{X}_i^n$ and  $\bm{Y}_i = \bm{Y}_i^1 + \bm{Y}_i^2 + ... + \bm{Y}_i^n$. For all the $n$ decomposed components, we set up $n$ models $f^1, f^2, ... f^n$ to individually learn the corresponding component. The final prediction is given by the reconstruction of different models' prediction.

    In our approach, we use Debauches (Db)-type mother wavelet to perform wavelet decomposition. The decomposition is performed for three times. An Approximate (A) component and a Detail (D) component are derived by low- and high-pass filtering in each step, resulting in three approximate components and three detail components, named $rA_1$, $rA_2$, $rA_3$, $rD_1$, $rD_2$, and $rD_3$. An example of the original the SWH data and decomposed six components are shown in Fig.~\ref{fig:Wavelet123}. The approximate components contain only the low-frequency component, and their high-frequency components are separated and moved into detail components at each decomposition step. 

\begin{figure}[h]
    \centering
    \includegraphics[width=0.47\textwidth]{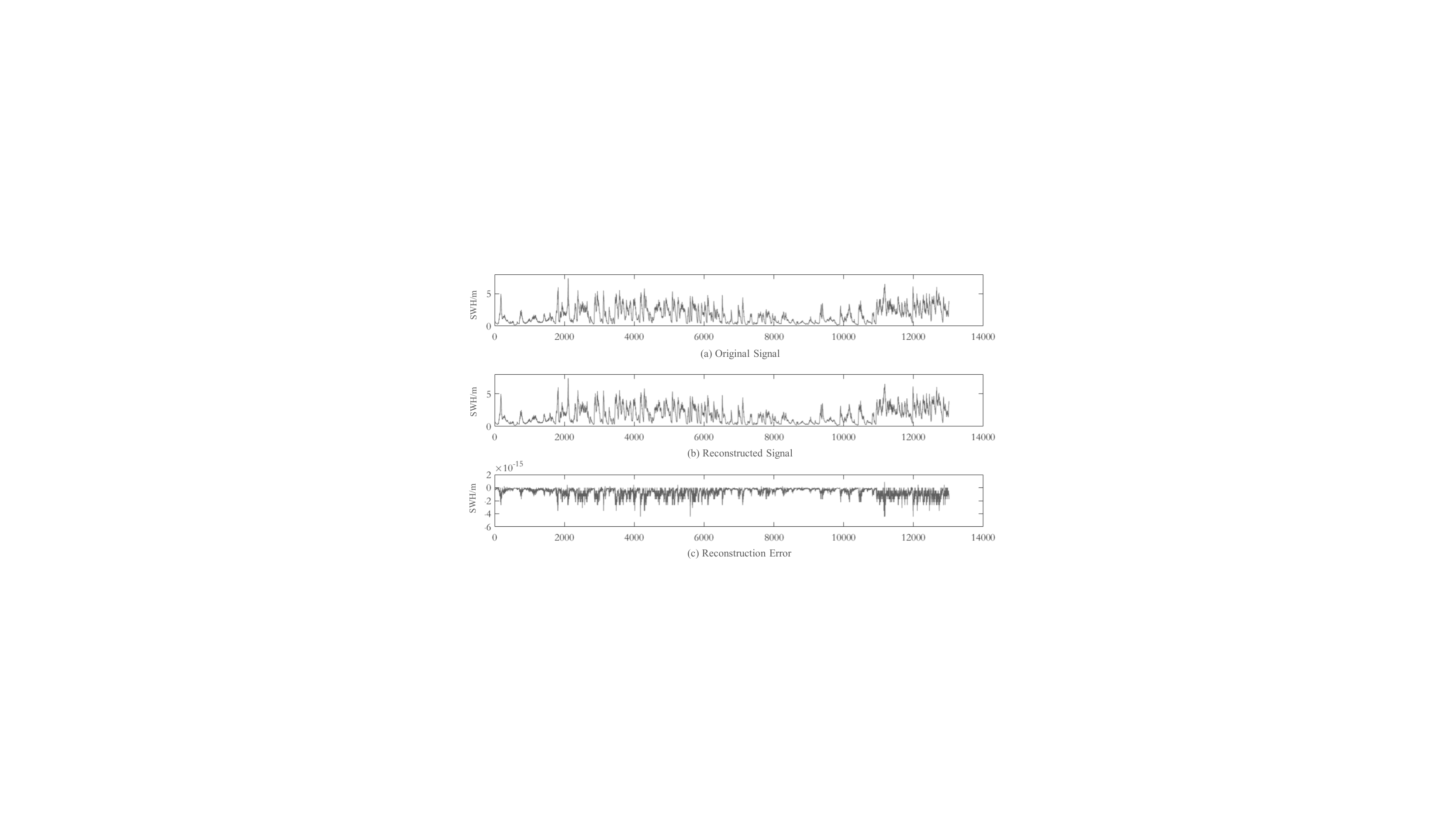}
    \caption{An example of SWH wavelet reconstruction and the reconstruction error.}
    \label{fig:WaveletReconstruction}
\end{figure}

\begin{table}[h]
    \centering
    \caption{The reconstruction error of different type of input}
    \begin{tabular}{ccc}
        \hline\hline
        Input              & Unit   & Reconstruction Error   \\ \hline
        Average wind speed & m/s    & $1.7764\times10^{-14}$ \\
        Maxim wind speed   & m/s    & $2.1316\times10^{-14}$ \\
        Wind Direction     & degree & $2.2737\times10^{-13}$ \\ \hline\hline
    \end{tabular}
    \label{tab:ReconstructionError}
\end{table}

    Note that the wavelet decomposition is reversible. Fig. \ref{fig:WaveletReconstruction} shows a comparison of original and reconstructed SWH data, and Table.\ref{tab:ReconstructionError} summarizes the reconstruction error of inputs data. From these result we can conclude that the error of both input data and target output SWH data is at a very low numerical level and can be ignored, showing the reliability of the wavelet transform approach.

    As shown in Fig.\ref{fig:WGNN}, among the six components, including $rA_1$, $rA_2$, $rA_3$, $rD_1$, $rD_2,$ and $rD_3$ derived from wavelet transform, we use $rD_1$, $rD_2$, $rD_3$, and $rA_3$ as the models input and output, i.e. $n=4$. Those four components are independently learned by four learning models, which is based on a Graph Neural Network (GNN) structure \cite{Wu2020Connecting}. Each model can only observe its own frequency band of input. The outputs of the four GNNs are subsequently reconstructed and form the final prediction outputs.
    
    The GNN is an effective learning model designed for multivariate time series prediction. It consists of a graph learning module, a graph convolution module, and a temporal convolution module. It also integrate a mix-hop propagation layer and a dilated inception layer \cite{Wu2020Connecting}. The loss function WGNN is:
    \begin{equation}
        \mathcal{L} =\frac{1}{N}\sum_{i=0}^N {||\sum_{k=1}^4 f^k(x_i) - y_i||_2^2}
    \end{equation}
    , where $f^1$, $f^2$, $f^3$, and $f^4$ are respectively for $rD_1$, $rD_2$, $rD_3$, and $rA_3$ derived from wavelet decomposition.

\section{Experiments}
\label{sec:experiments}
\subsection{Experiment Setup}
    We implemented our proposed approach by MATLAB and Tensorflow. The WGNN consists of five spatial and five temporal convolution modules. The dilatation is set to 2. We train the WGNN model for 30 epochs with a batchsize of 4.
    
    The data comes from the a buoy deployed by the Chinese Fujian Wave Forecast Station in the middle of the Taiwan Strait, whose latitude and longitude are 119.30°E and 24.48°N respectively. The dataset is collected from July 1, 2016 to December 31, 2017, and consist of a total of 13,076 records with the time resolution of one hour, which is normalized to [0,1]. The train-validate-test split ratio is 6:2:2.

\subsection{Comparisons and Results}
    To validate the effeteness of proposed WGNN, we compare its performance with the following three types baseline: 
    \begin{itemize}
        \item \textbf{Numerical model}. For the numerical model, the widely used WAVEWATCH III (WW3)\cite{Tolman2002Development} forecasting model is involved.
        \item \textbf{Machine learning models}. Machine learning models involve four types of classical machine learning algorithms ANN, SVM, RNN, and LSTM.
        \item \textbf{Deep learning models}. For deep learning comparisons, SAE-BP \cite{Liu2017Flood}, Long- and Short-term Time-series Network (LSTNet) \cite{Lai2018Modeling}, Temporal Pattern Attention LSTM (TPA-LSTM) \cite{Shih2019Temporal} are involved. SAE-BP \cite{Liu2017Flood} is a deep learning approach by integrating stacked auto-encoders (SAE) and back-propagation neural networks (BPNN) for the prediction of stream flow. LSTNet uses the CNN and the RNN to extract short-term local dependency patterns and discover long-term patterns for time series trends. TPA-LSTM combines temporal pattern attention mechanism with LSTM.
    \end{itemize}
    
    We use the Mean Square Error (MSE) and R$^2$ as the evaluation metrics, whose definitions are respectively showed in Eq.\ref{Eq:MSE} and Eq.\ref{Eq:R2},
    \begin{equation}
        MSE = \frac{1}{N}\sum_{i=0}^N{(\hat{y_i} - y_i)^2}
        \label{Eq:MSE}
    \end{equation}
    \begin{equation}
        R^2 = 1-\frac{\sum_{i=0}^N{(\hat{y_i} - y_i)^2}} {\sum_{i=0}^N{(\overline{y} - y_i)^2}}
        \label{Eq:R2}
    \end{equation}
    , where $\hat{y_i}$ is the output of SWH prediction model, $y_i$ is the ground truth data, and $\overline{y}$ is the average value of ground truth data.

    \begin{table}[h]
        \centering
        \caption{The performance comparison of different models.}
            \begin{tabular}{cccc}	
            \hline\hline
            Model                & Reference              & MSE $\downarrow$& R$^2\uparrow$   \\ \hline
            WW3                  & \cite{Tolman2002Development}& 0.2960     & 0.7928          \\
            SVM                  & -                      & 0.2176          & 0.8442          \\
            ANN                  & -                      & 0.2730          & 0.8229          \\
            RNN                  & -                      & 0.7204          & 0.5333          \\
            LSTM                 & -                      & 0.2138          & 0.8469          \\
            SAE-BP               & \cite{Liu2017Flood}    & 0.2539          & 0.8357          \\
            LSTNet               & \cite{Lai2018Modeling} & 0.1458          & 0.8984          \\
            TPA-LSTM             & \cite{Shih2019Temporal}& 0.1882          & 0.8780          \\
            GNN                  & Ours, no wavelet       & 0.1420          & 0.9135          \\
            \textbf{WGNN}        & \textbf{Ours}          & \textbf{0.1187} & \textbf{0.9341} \\ \hline\hline
    
            \end{tabular}
        \label{tab:Comparisons}
    \end{table}
    
    \begin{figure*}[h]
        \centering
        \includegraphics[width=0.9\textwidth]{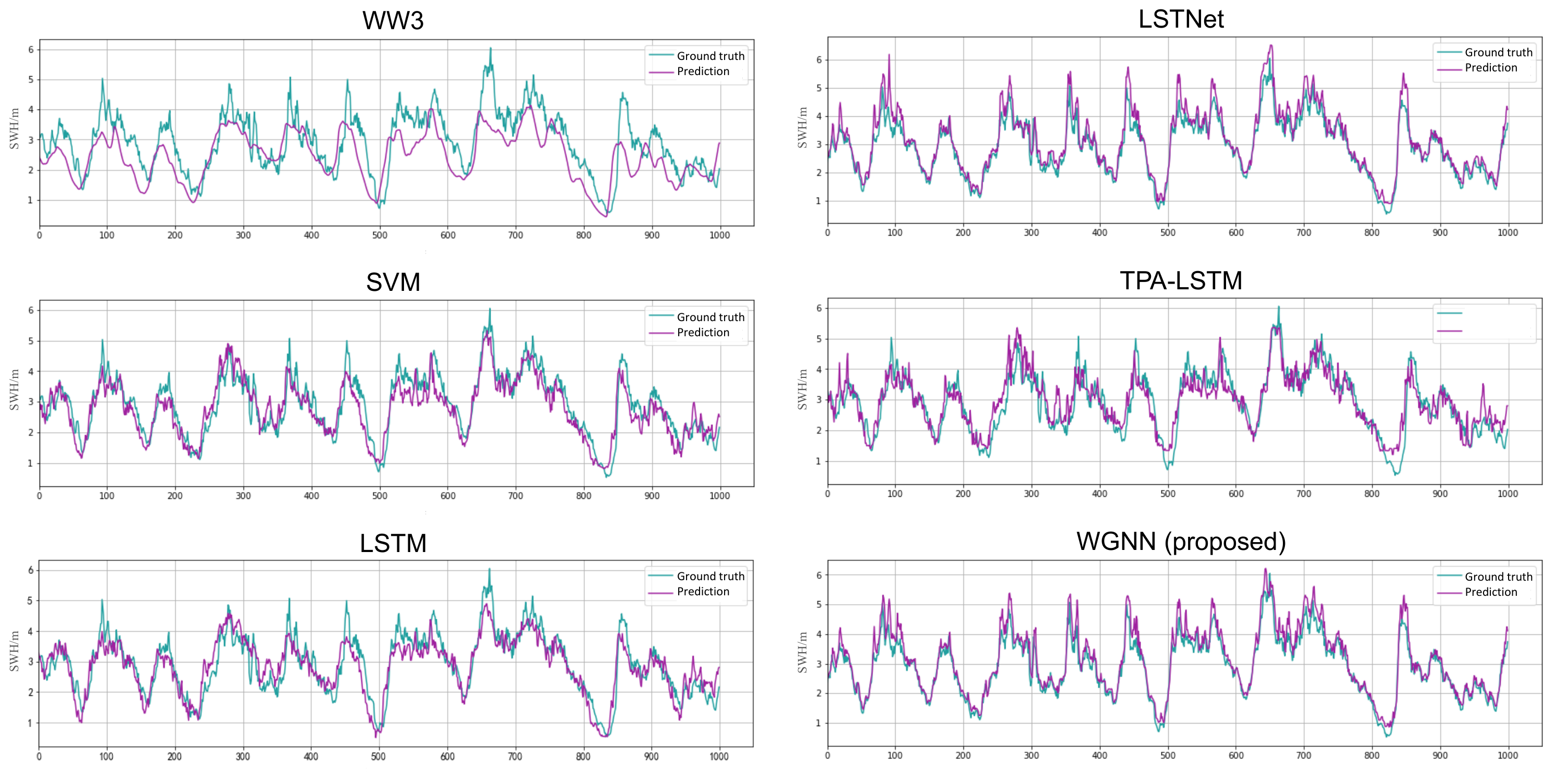}
        \caption{The predicted results of baselines and proposed WGNN.}
        \label{fig:PredictedWaves}
    \end{figure*}

    Table \ref{tab:Comparisons} presents the experiment results. Since numerical models do not have the learning ability, the WW3 model performed the worst. Machine learning-based models SVM and ANN yield better results, but they still lack the capability of learning time dependencies. The RNN fails due to the gradient vanishing problem, while LSTM successfully solves this problem and achieves a more stable training procedure and better accuracy. Compared to all the involved simple baselines and existing methods \cite{Liu2017Flood, Lai2018Modeling, Shih2019Temporal}, the proposed WGNN achieved the lowest MSE and highest R$^2$. In addition, GNN achieved the second best performance, showing that the graph-based learning scheme can effectively capture and learn from the relationships between different inputs.
    
    In Fig.\ref{fig:PredictedWaves}, we show the comparisons of ground truth SWH and predictions of different models. The numerical method fits the approximate contour of the target, but it fails to predict several extreme situations. Other learning-based ``soft computing'' models slightly alleviated this problem, but they did not completely solve it. In contrast, since the proposed WGNN-based method explicitly decomposed the data in the frequency domain, it can separately learn and effectively handle both normal and extreme situations. 
    
\section{Conclusion}
\label{sec:conclusion}
In this paper, a Wavelet Graph Neural Network (WGNN) is proposed for Significant Wave Height (SWH) prediction. By integrating wavelet transformation and graph neural network, our model experimentally outperforms various prediction models, including numerical models, machine learning models, and several deep learning models. We also demonstrate the effectiveness of wavelet decomposition and reconstruction. Compared to the vanilla GNN, wavelet transform yields a 16.4\% MSE reduction. It also demonstrates that deep learning can cooperate well with wavelet decomposition.

\bibliographystyle{IEEEtran}
\bibliography{SignificantWaveHight}
\end{document}